\documentclass[10pt,twocolumn,letterpaper]{article}
\newif\ifcomments
\commentsfalse
\commentstrue
\usepackage[nocompress]{cite}
\usepackage{cvpr}
\usepackage{times}
\usepackage{epsfig}
\usepackage{graphicx}
\usepackage{amsmath}
\usepackage{amssymb}
\usepackage{wrapfig}
\usepackage{multirow}
\usepackage{booktabs}
\usepackage{bm}

\renewcommand{\paragraph}[1]{\vspace{0.25ex}\noindent\textbf{#1}}

\usepackage[pagebackref=true,breaklinks=true,letterpaper=true,colorlinks,bookmarks=false]{hyperref}

\cvprfinalcopy 

\ifcvprfinal\fi
\begin{document}

\ifcomments
  \newcommand{\comments}[1]{#1}
\else
  \newcommand{\comments}[1]{}
\fi

\newcommand{\name}{ConDenseNet}

\newcommand{\kqw}[1]{\comments{\textcolor{blue}{[Kilian: #1]}}}
\newcommand{\lvdm}[1]{\comments{\textcolor{green}{[Laurens: #1]}}}
\newcommand{\todo}[1]{\comments{\textcolor{red}{[Todo: #1]}}}

\title{\condense: An Efficient DenseNet using \methodnamecap{}s}

\author{Gao Huang$\thanks{Both authors contributed equally.}$\\
Cornell University\\
{\tt\small gh349@cornell.edu}
\and
Shichen Liu$^{*}$\\
Tsinghua University\\
{\tt\small liushichen95@gmail.com}
\and
Laurens van der Maaten\\
Facebook AI Research\\
{\tt\small lvdmaaten@fb.com}
\and
Kilian Q. Weinberger\\
Cornell University\\
{\tt\small kqw4@cornell.edu}
}  


\newcommand{\modelname}{DenseNet-{Lite}}
\newcommand{\methodname}{learned group convolution}
\newcommand{\methodnamecap}{Learned Group Convolution}
\newcommand{\methodnameshort}{LGC}
\newcommand{\condense}{CondenseNet}

\newcommand{\methodblock}{dense block}
\newcommand{\methodblockcap}{Dense Block}

\newcommand{\regmethodname}{feature drop}
\newcommand{\regmethodnamecap}{Feature Drop}

\newcommand{\stepsizename}{growth rate}

\newcommand{\red}[1]{\textcolor{red}{#1}}
\newcommand{\blue}[1]{\textcolor{blue}{#1}}

\newcommand{\conv}[2]{$
\left[
\begin{array}{ll}
{1}\!\times\! {1} \text{ L-Conv} \\
{3}\!\times\! {3} \text{ G-Conv} \\
\end{array}
\right] \!\times\! {#1} \quad (k\!=\!#2)$}

\newcommand{\group}[4]{\multicolumn{1}{c#4}{$\text{G = #1, C = #2, G = #3}$}}

\newcommand{\cross}[1]{$#1 \!\times\! #1$}

\newcommand{\feati}{x_i}
\newcommand{\clsfeati}{y_i}
\newcommand{\featk}{x_k}
\newcommand{\clsfeatk}{y_k}
\newcommand{\loss}{L}
\newcommand{\featL}{x_L}
\newcommand{\clsfeat}{y}
\newcommand{\anyxs}{\ensuremath{\mathbf{x}}}
\newcommand{\anyys}{\ensuremath{\mathbf{y}}}

\newcommand{\bF}{\ensuremath{\mathbf{F}}}
\newcommand{\bM}{\ensuremath{\mathbf{M}}}

\newcommand{\bx}{\ensuremath{\mathbf{x}}}

\newcommand{\sourcexs}{\ensuremath{\mathbf{x^\mathcal{S}}}}
\newcommand{\sourceys}{\ensuremath{\mathbf{y^\mathcal{S}}}}

\newcommand{\targetxs}{\ensuremath{\mathbf{x^\mathcal{T}}}}
\newcommand{\targetys}{\ensuremath{\mathbf{y^\mathcal{T}}}}
\newcommand{\pseudotargetys}{\ensuremath{\mathbf{\hat{y}^\mathcal{T}}}}

\maketitle

\begin{abstract}
Deep neural networks are increasingly used on mobile devices, where computational resources are limited. In this paper we develop \condense{}, a novel network architecture with unprecedented efficiency.
It combines dense connectivity with a novel module called \methodname{}. The dense connectivity facilitates feature re-use in the network, whereas \methodname{}s remove connections between layers for which this feature re-use is superfluous. At test time, our model can be implemented using standard group convolutions, allowing for efficient computation in practice. Our experiments show that \condense{}s are far more efficient than state-of-the-art compact convolutional networks such as ShuffleNets.

\end{abstract}


\vspace{-2ex}
\section{Introduction}
The high accuracy of convolutional networks (CNNs) in visual recognition tasks, such as image classification \cite{vgg,resnet,huang2017densely}, has fueled the desire to deploy these networks on platforms with limited computational resources, \emph{e.g.}, in robotics, self-driving cars, and on mobile devices. Unfortunately, the most accurate deep CNNs, such as the winners of the ImageNet \cite{deng2009imagenet} and COCO \cite{lin2014microsoft} challenges, were designed for scenarios in which computational resources are abundant. As a result, these models cannot be used to perform real-time inference on low-compute devices.

This problem has fueled development of computationally efficient CNNs that, \emph{e.g.}, prune redundant connections \cite{lecun1989optimal,hassibi1993optimal,li2016pruning,liu2017learning,han2015deep}, use low-precision or quantized weights \cite{hubara2016binarized,rastegari2016xnor,chen2015compressing}, or use more efficient network architectures \cite{iandola2016squeezenet,resnet,huang2017densely,chollet2016xception,howard2017mobilenets,zhang2017shufflenet}. These efforts have lead to substantial improvements: to achieve comparable accuracy as VGG~\cite{vgg} on ImageNet, ResNets \cite{resnet} reduce the amount of computation by a factor $5 \times$, DenseNets \cite{huang2017densely} by a factor of $10 \times$, and MobileNets \cite{howard2017mobilenets} and ShuffleNets \cite{zhang2017shufflenet} by a factor of $25\times$. A typical set-up for deep learning on mobile devices is one where CNNs are trained on multi-GPU machines but deployed on devices with limited compute. Therefore, a good network architecture allows for fast parallelization during training, but is compact at test-time.


Recent work~\cite{chen2015compressing,stochastic} shows that there is a lot of redundancy in CNNs. The layer-by-layer connectivity pattern forces networks to replicate features from earlier layers throughout the network. The DenseNet architecture \cite{huang2017densely} alleviates the need for feature replication by directly connecting each layer with all layers before it, which induces feature re-use.
Although more efficient, we hypothesize that dense connectivity introduces redundancies when early features are not needed in later layers. We propose a novel method to prune such redundant connections between layers and then introduce a more efficient architecture. In contrast to prior pruning methods, our approach learns a sparsified network automatically during the training process, and  produces a regular connectivity pattern that can be implemented efficiently using group convolutions. Specifically, we split the filters of a layer into multiple groups, and gradually remove the connections to less important features \emph{per group} during training. Importantly, the groups of incoming features are not predefined, but \emph{learned}. The resulting model, named \condense,  can be trained efficiently on GPUs, and has high inference speed on mobile devices.

Our image-classification experiments show that \condense{}s consistently outperform alternative network architectures. Compared to DenseNets, \condense{}s use only $1/10$ of the computation at comparable accuracy levels. On the ImageNet dataset \cite{deng2009imagenet}, a \condense{} with 275 million FLOPs\footnote{Throughout the paper, \emph{FLOPs} refers to the number of multiplication-addition operations.} achieved a 29\% top-1 error, which is comparable to the error of a MobileNet that requires twice as much compute.




\vspace{-1ex}
\section{Related Work and Background}
\vspace{-1ex}

%
%

We first review related work on model compression and efficient network architectures, which inspire our work. Next, we review the DenseNets and group convolutions that form the basis for \condense.

\vspace{-1ex}
\subsection{Related Work}
\vspace{-1ex}
\noindent \textbf{Weights pruning and quantization.}
\condense{}s are closely related to approaches that improve the inference efficiency of (convolutional) networks via weight pruning~\cite{lecun1989optimal,hassibi1993optimal,li2016pruning,liu2017learning,he2017channel} and/or weight quantization~\cite{hubara2016binarized,rastegari2016xnor}. These approaches are effective because deep networks often have a substantial number of redundant weights that can be pruned or quantized without sacrificing (and sometimes even improving) accuracy. For convolutional networks, different pruning techniques may lead to different levels of granularity~\cite{Mao}. Fine-grained pruning, \emph{e.g.}, independent weight pruning \cite{lecun1989optimal,han2015learning}, generally achieves a high degree of sparsity. However, it requires storing a large number of indices, and relies on special hardware/software accelerators. In contrast, coarse-grained pruning methods such as filter-level pruning \cite{li2016pruning,alvarez2016learning,liu2017learning,he2017channel} achieve a lower degree of sparsity, but the resulting networks are much more regular, which facilitates efficient implementations.

\condense{}s also rely on a pruning technique, but differ from prior approaches in two main ways: First, the weight pruning is initiated in the early stages of training, which is substantially more effective and efficient than using $L_1$ regularization throughout. Second, \condense{}s have a higher degree of sparsity than filter-level pruning, yet generate highly efficient group convolution---reaching a sweet spot between sparsity and regularity.

%

\noindent \textbf{Efficient network architectures.} A range of recent studies has explored efficient convolutional networks that can be trained end-to-end  \cite{huang2017densely,zhang2017interleaved,howard2017mobilenets,zhang2017shufflenet,zoph2017learning,iandola2016squeezenet,zhaodeep}.
Three prominent examples of networks that are sufficiently efficient to be deployed on mobile devices are MobileNet \cite{howard2017mobilenets}, ShuffleNet \cite{zhang2017shufflenet}, and Neural Architecture Search (NAS) networks \cite{zoph2017learning}. All these networks use depth-wise separable convolutions, which greatly reduce computational requirements without significantly reducing accuracy. A practical downside of these networks is depth-wise separable convolutions are not (yet) efficiently implemented in most deep-learning platforms. By contrast, \condense{} uses the well-supported group convolution operation~\cite{alexnet}, leading to better computational efficiency in practice.

\noindent \textbf{Architecture-agnostic efficient inference}
has also been explored by several  prior studies. For example, knowledge distillation \cite{bucilu2006model,hinton2015distilling} trains small ``student'' networks to reproduce the output of large ``teacher'' networks to reduce test-time costs. Dynamic inference methods \cite{bolukbasi2017adaptive,graves2016adaptive,figurnov2016spatially,huang2018multi} adapt the inference to each specific test example, skipping units or even entire layers to reduce computation. We do not explore such approaches here, but believe they can be used in conjunction with \condense{}s.

\begin{figure}
  \centering
  \includegraphics[width=0.38 \textwidth]{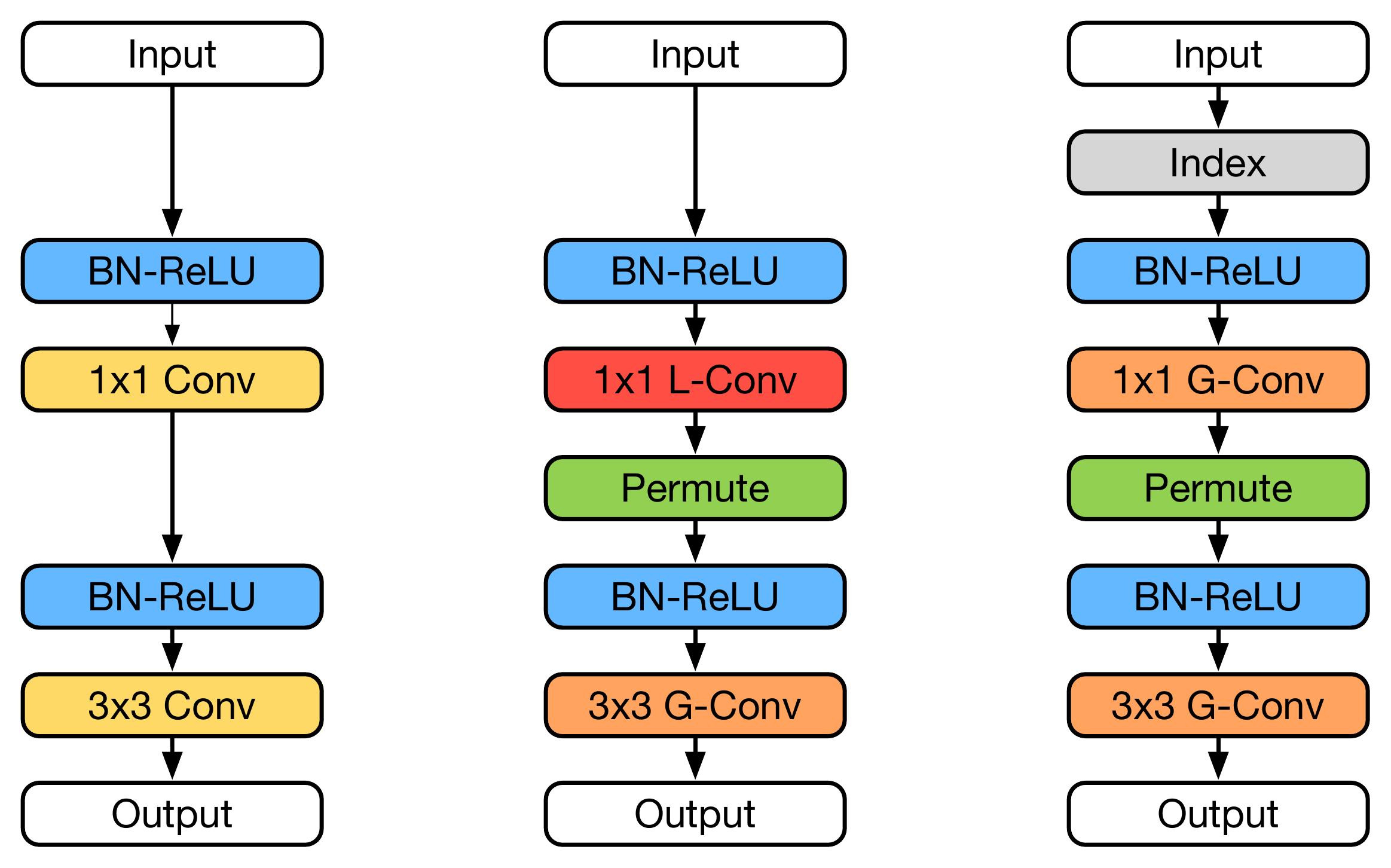}\\
  \caption{The transformations within a \emph{layer} in DenseNets \emph{(left)}, and   \condense{}s at training time \emph{(middle)} and at test time \emph{(right)}.  The \emph{Index} and \emph{Permute} operations are explained in Section~\ref{subsec:LGC} and \ref{subsec:cifar}, respectively. (L-Conv: learned group convolution; G-Conv: group convolution) }
  \vspace{-2ex}
  \label{fig:structure_compare}
\end{figure}

\begin{figure}
  \centering
  \includegraphics[width=0.45 \textwidth]{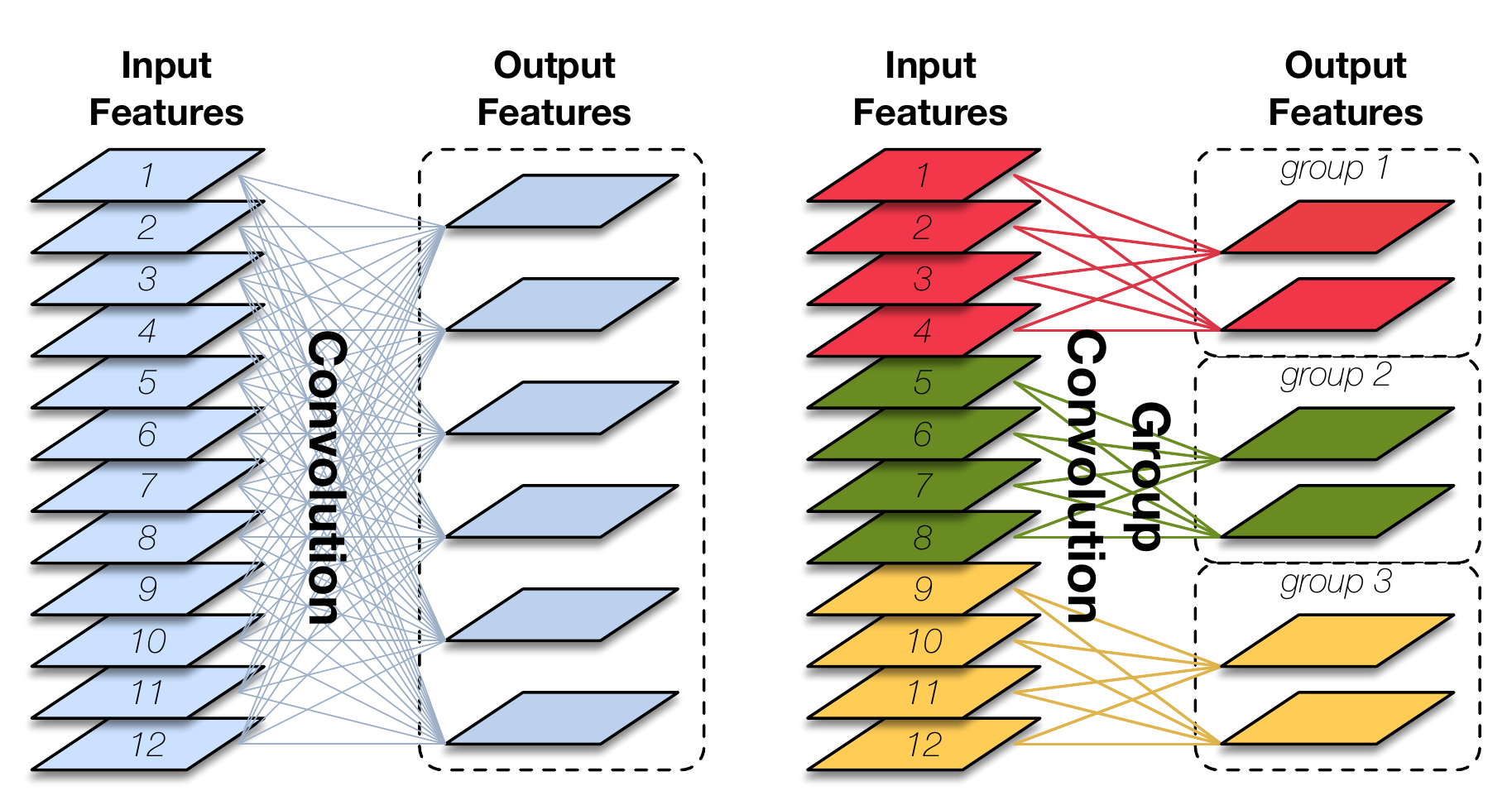}\\
  \caption{Standard convolution \emph{(left)} and group convolution \emph{(right)}. The latter enforces a sparsity pattern by partitioning the inputs (and outputs) into disjoint groups.  }
  \vspace{-2ex}
  \label{fig:group-conv}
\end{figure}


\begin{figure*}[t]
  \centering
  \includegraphics[width=1.0 \textwidth]{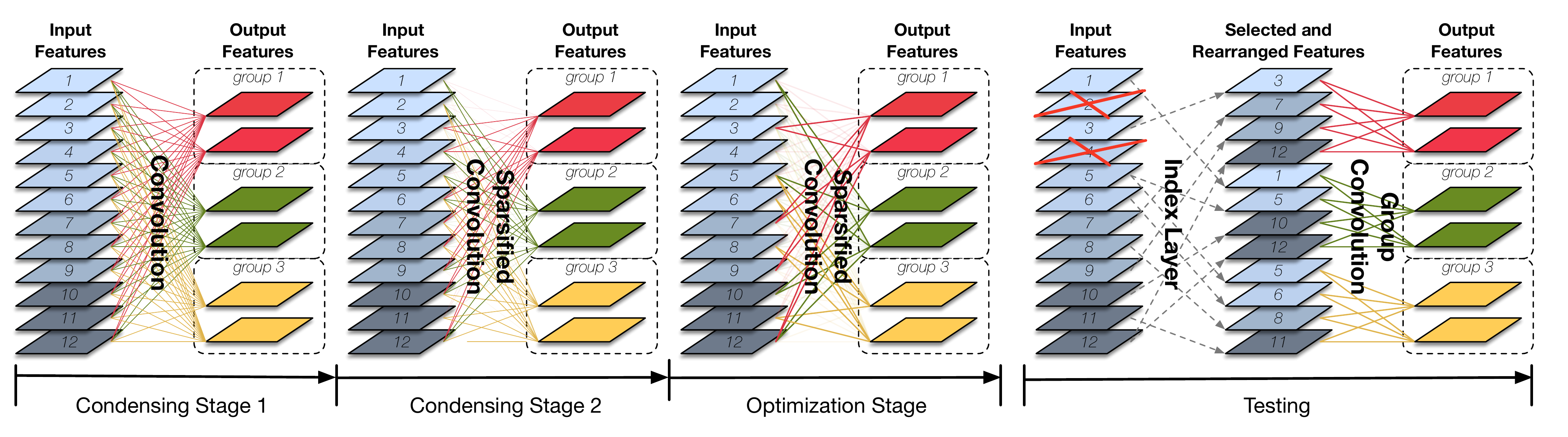}\\
  \caption{Illustration of \methodname{}s with $G \!=\! 3$ groups and a condensation factor of $C \!=\! 3$. During training a fraction of $(C\!-\!1)/C$ connections are removed after each of the $C-1$ condensing stages. Filters from the same group use the same set of features, and during test-time the \emph{index layer} rearranges the features to allow the  resulting model to be implemented as standard group convolutions.}\label{fig:learned_conv}
  \vspace{-2ex}
\end{figure*}

\vspace{-1ex}
\subsection{DenseNet}
\vspace{-1ex}
Densely connected networks (DenseNets; \cite{huang2017densely}) consist of multiple \emph{dense blocks}, each of which consists of multiple \emph{layers}. Each layer produces $k$ features, where $k$ is referred to as the \emph{growth rate} of the network. The distinguishing property of DenseNets is that the input of each  layer is a concatenation of all feature maps generated by \emph{all} preceding layers within the same dense block. 
Each  layer performs a sequence of consecutive  transformations, as shown in the left part of \figurename~\ref{fig:structure_compare}.  The first transformation (\emph{BN-ReLU}, blue) is a composition of {batch normalization} \cite{batch-norm} and {rectified linear units} \cite{nair2010rectified}.
The first convolutional layer in the sequence reduces the number of channels to save computational cost by using the $1\!\times\! 1$ filters. The output is followed by another BN-ReLU transformation and is then reduced to the final $k$ output features through a $3\!\times \!3$ convolution.

\vspace{-1ex}
\subsection{Group Convolution}
\vspace{-1ex}
\label{sec:group-conv}
Group convolution is a special case of a sparsely connected convolution, as illustrated in \figurename~\ref{fig:group-conv}. It was first used in the AlexNet architecture \cite{alexnet}, and has more recently been popularized by their successful application in ResNeXt \cite{resnext}. Standard convolutional layers (left illustration in \figurename~\ref{fig:group-conv}) generate $O$ output features by applying a convolutional filter (one per output) over \emph{all} $R$ input features, leading to a computational cost of $R\times O$. In comparison, group convolution (right illustration) reduces this computational cost by partitioning the input features into $G$ mutually exclusive groups, each producing its own outputs---reducing the computational cost by a factor $G$ to $\frac{R\times O}{G}$.

\section{\condense s}
\label{sec:method}

Group convolution works well with many deep neural network architectures~\cite{resnext,zhang2017shufflenet,zhang2017interleaved} that are connected in a layer-by-layer fashion.
For dense architectures group convolution can be used in the $3\!\times \!3$ convolutional layer (see Figure~\ref{fig:structure_compare}, left). However, preliminary experiments show that a na\"ive adaptation of group convolutions in the $1\!\times \!1$ convolutional layer leads to drastic reductions in accuracy. We surmise that this is caused by the fact that the inputs to the $1\!\times \!1$ convolutional layer are concatenations of feature maps generated by preceding layers. Therefore, they differ in two ways from typical inputs to convolutional layers: 1. they have an intrinsic order; and 2. they are far more diverse.
The hard assignment of these features to disjoint groups hinders effective re-use of features in the network.
Experiments in which we randomly permute input feature maps in each layer before performing the group convolution show that this reduces the negative impact on accuracy --- but even with the random permutation, group convolution in the $1\!\times \!1$ convolutional layer makes DenseNets less accurate than for example smaller DenseNets with equivalent computational cost.

It is shown in \cite{huang2017densely} that making early features available as inputs to later layers is important for efficient feature re-use.
Although not all prior features are needed at every subsequent layer, it is hard to predict which features should be utilized at what point.
To address this problem, we develop an approach that \emph{learns} the input feature groupings automatically during training. Learning the group structure allows each filter group to select its own set of most relevant inputs. Further, we allow multiple groups to share input features and also allow features to be ignored by all groups. Note that in a DenseNEt, even if an input feature is ignored by all groups in a specific layer, it can still be utilized by some groups At different layers.
To differentiate it from regular group convolutions, we refer to our approach as \emph{\methodname}.

\subsection{\methodnamecap}
\label{subsec:LGC}

\begin{figure}[t]
	\centering
	\includegraphics[width=0.5 \textwidth]{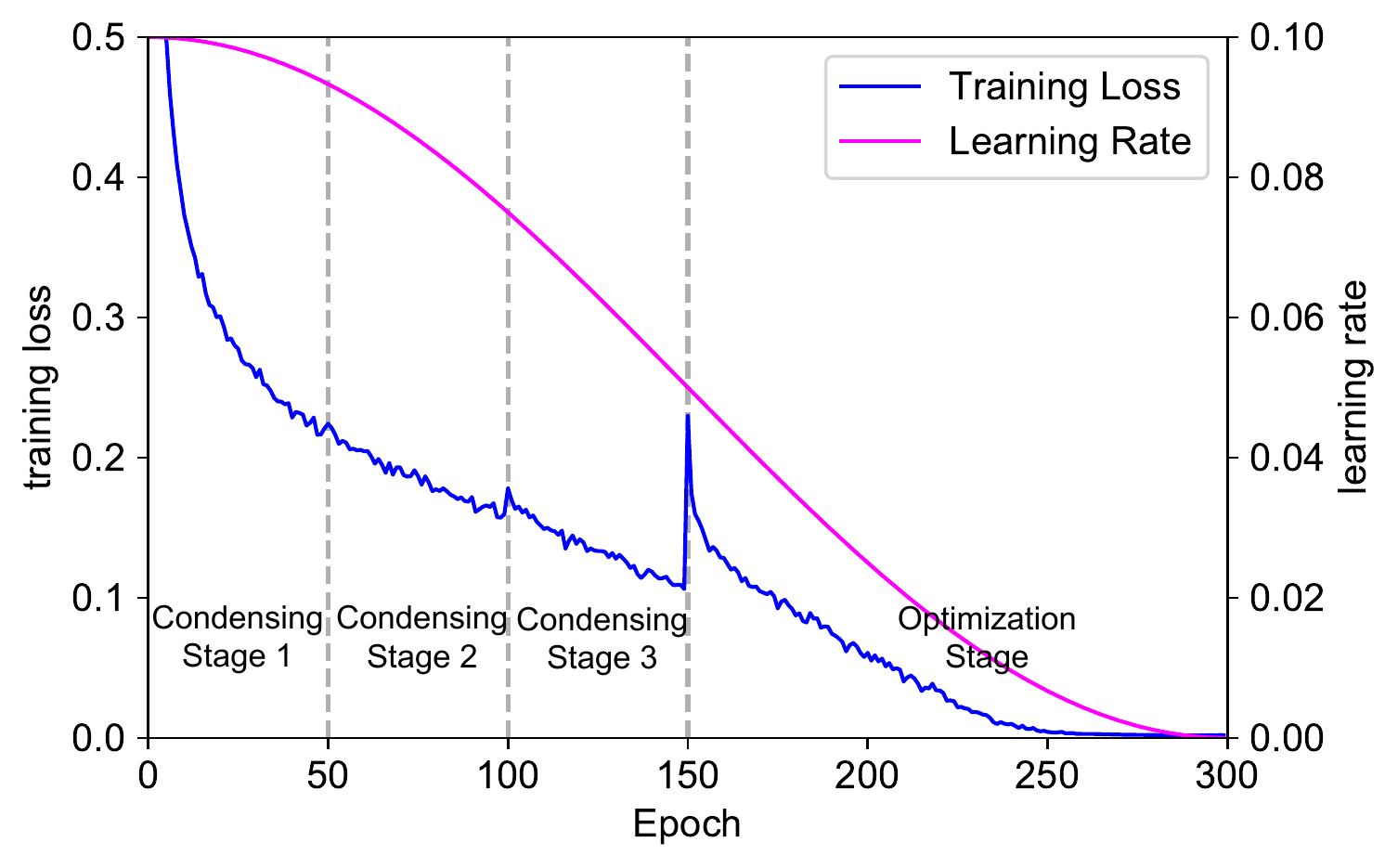}
	\caption{The cosine shape learning rate and a typical training loss curve with a condensation factor of $C \!=\! 4$.}
	\label{fig:learning-rate}
    \vspace{-2ex}
\end{figure}

We learn group convolutions through a multi-stage process, illustrated in
\figurename{}s~\ref{fig:learned_conv} and \ref{fig:learning-rate}. The first half of the training iterations comprises of \emph{condensing} stages. Here,  we repeatedly train the network with sparsity inducing regularization for a fixed number of iterations and subsequently prune away unimportant filters with low magnitude weights. The second half of the training consists of the \emph{optimization} stage, in which we learn the filters after the groupings are fixed.
When performing the pruning, we ensure that filters from the same group share the same sparsity pattern. As a result, the sparsified layer can be implemented using a standard group convolution once training is completed (\emph{testing} stage). Because group convolutions are efficiently implemented by many deep-learning libraries, this leads to high computational savings both in theory and in practice. We present details on our approach below.

\paragraph{Filter Groups.} We start with a standard convolution of which filter weights form a 4D tensor of size $O\!\times\! R \!\times\! W \!\times\! H$, where $O$, $R$, $W$, and $H$ denote the number of output channels, the number of input channels, and the width and the height of the filter kernels, respectively. As we are focusing on the $1\times 1$ convolutional layer in DenseNets, the 4D tensor reduces to an $O\!\times\! R$ matrix $\bF$. We consider the simplified case in this paper. But our procedure can readily be used with larger convolutional kernels. Before training, we first split the filters (or, equivalently, the output features) into $G$ groups of equal size. We denote the filter weights for these groups by $\bF^1,\ldots,\bF^G$; each $\bF^g$ has size $\frac{O}{G}\!\times\! R$ and $\bF^g_{ij}$ corresponds to the weight of the $j$th input for the $i$th output within group $g$. Because the output features do not have an implicit ordering, this random grouping does not negatively affect the quality of the layer.

\paragraph{Condensation Criterion.}
During the training process we gradually screen out  subsets of less important input features for each group. The importance of the $j$th incoming feature map for the filter group $g$ is evaluated by the averaged absolute value of weights between them across all outputs within the group, \emph{i.e.}, by $\sum_{i=1}^{{O}/{G}} |\bF^g_{i,j}|$. In other words, we remove columns in $\bF^g$ (by zeroing them out) if their $L_1$-norm is small compared to the $L_1$-norm of other columns. This results in a convolutional layer that is structurally sparse: filters from the same group always receive the same set of features as input.

\paragraph{Group Lasso.} To reduce the negative effects on accuracy introduced by weight pruning,  $L_1$ regularization is commonly used to induce sparsity \cite{li2016pruning,liu2017learning}. In \condense s, we encourage convolutional filters from the same group to use the same subset of incoming features, \emph{i.e.}, we induce group-level sparsity instead. To this end, we use the following group-lasso regularizer \cite{yuan2006model} during training:
\begin{equation}
\label{eq:group-lasso}
\sum\nolimits_{g=1}^G\sum\nolimits_{j=1}^{R}\sqrt{\sum\nolimits_{i=1}^{O/G} {\bF^{g}_{i,j}}^2}. \nonumber
\end{equation}
The group-lasso regularizer simultaneously pushes all the elements of a column of $\bF^g$ to zero, because the term in the square root is dominated by the largest elements in that column. This induces the group-level sparsity we aim for.

\paragraph{Condensation Factor.} In addition to the fact that \methodname{}s are able to automatically discover good connectivity patterns, they are also more flexible than standard group convolutions.
In particular, the proportion of feature maps used by a group does not necessarily need to be $\frac{1}{G}$. We define a \emph{condensation factor} $C$, which may differ from $G$, and allow each group to select $\left\lfloor{\frac{R}{C}}\right\rfloor$ of inputs.

\paragraph{Condensation Procedure.}
In contrast to approaches that prune weights in pre-trained networks, our weight pruning process is integrated into the training procedure. As illustrated in \figurename~\ref{fig:learned_conv} (which uses $C\!=\!3$),
at the end of each  $C\!-\!1$ \emph{condensing} stages we prune $\frac{1}{C}$ of the filter weights.
By the end of training, only $\frac{1}{C}$ of the weights remain in each filter group.  In all our experiments we set the number of training epochs of the condensing stages to $\frac{M}{2(C\!-\!1)}$, where $M$ denotes the total number of training epochs---such that the first half of the training epochs is used for condensing.
In the second half of the training process, the \emph{Optimization} stage,  we train the sparsified model.\footnote{
In our implementation of the \emph{training} procedure we do not actually remove the pruned weights, but instead mask the filter $\bF$ by a binary tensor $\bM$ of the same size using an element-wise product.
The mask is initialized with only ones, and elements corresponding to pruned weights are set to zero. This implementation via masking is more efficient on GPUs, as it does not require sparse matrix operations. In practice, the pruning hardly increases the wall time needed to perform a forward-backward pass during training.}



\paragraph{Learning rate.} We adopt the \emph{cosine shape learning rate} schedule of Loshchilov et al.~\cite{loshchilov2017sgdr}, which smoothly anneals the learning rate, and usually leads to improved accuracy~\cite{huang2017snapshot,zoph2017learning}. \figurename~\ref{fig:learning-rate} visualizes the learning rate as a function of training epoch (in magenta), and the corresponding training loss (blue curve) of a \condense{} trained on the CIFAR-10 dataset \cite{cifar}. The abrupt increase in the loss at epoch 150 is causes by the final condensation operation, which removes half of the remaining weights. However, the plot shows that the model gradually recovers from this pruning step in the optimization stage.

\paragraph{Index Layer.} After training we remove the pruned weights and convert the sparsified model into a network with a regular connectivity pattern that can be efficiently deployed on devices with limited computational power. For this reason we introduce an \emph{index layer} that implements the feature selection and rearrangement operation (see
\figurename~\ref{fig:learned_conv}, right).
The convolutional filters in the output of the index layer are rearranged
to be amenable to existing (and highly optimized) implementations of regular group convolution. Figure~\ref{fig:structure_compare} shows the transformations of the \condense{} layers during training (middle) and during testing (right). During training the $1\times 1$ convolution is a learned group convolution (L-Conv), but during testing, with the help of the \emph{index} layer, it becomes a standard group convolution (G-Conv).

\begin{figure}
  \centering
  \includegraphics[width=0.5 \textwidth]{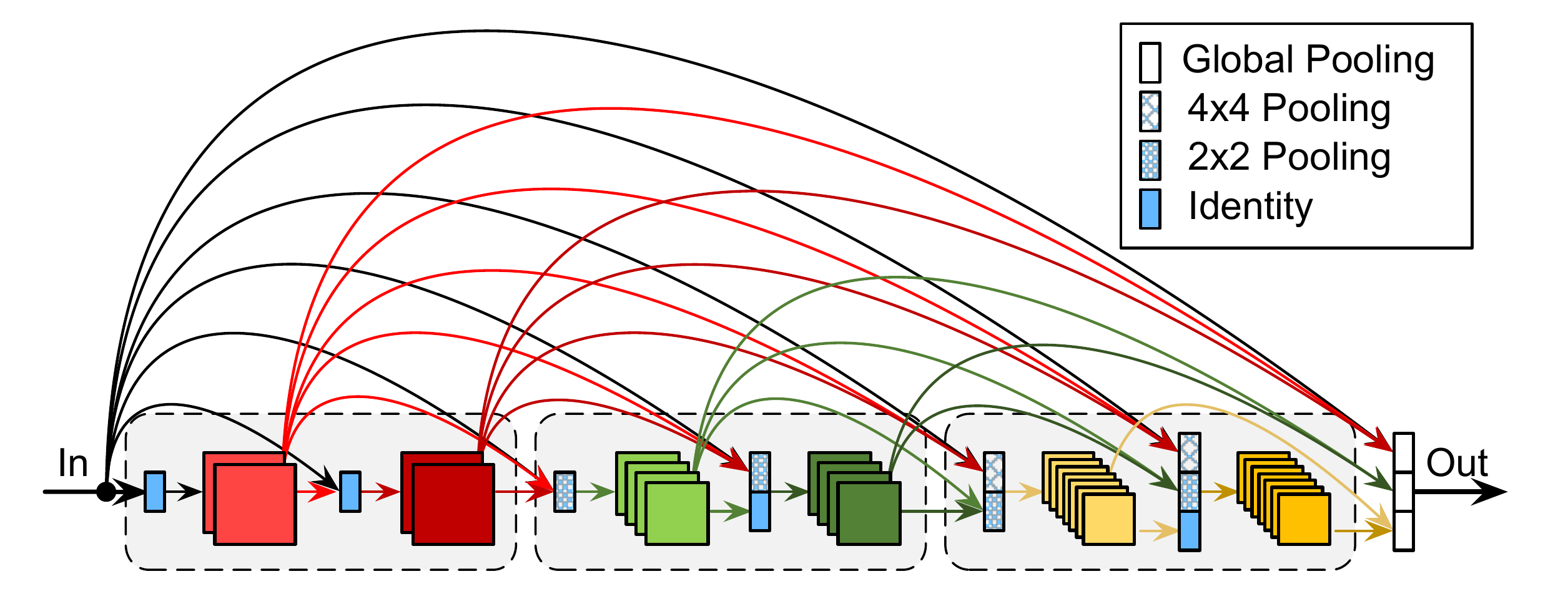}\\
  \caption{The proposed DenseNet variant. It differs from the original DenseNet in two ways: (1) layers with different resolution feature maps are also directly connected; (2) the growth rate doubles whenever the feature map size shrinks (far more features are generated in the third, yellow, dense block than in the first).}\label{fig:fully_densenet}
\vspace{-2ex}
\end{figure}


\subsection{Architecture Design}
\vspace{-1ex}
In addition to the use of \methodname{}s introduced above, we make two changes to the regular DenseNet architecture. These changes are designed to further simplify the architecture and improve its computational efficiency. \figurename~\ref{fig:fully_densenet} illustrates the two changes that we made to the DenseNet architecture.

\paragraph{Exponentially increasing growth rate.}
The original DenseNet design adds $k$ new feature maps at each layer, where $k$ is a constant referred to as the \emph{growth rate}. As shown in \cite{huang2017densely}, deeper layers in a DenseNet tend to rely on high-level features more than on low-level features. This motivates us to improve the network by strengthening \emph{short-range} connections.
We found that this can be achieved by gradually increasing the growth rate as the depth grows. This increases the proportion of features coming from later layers relative to those from earlier layers. For simplicity, we set the growth rate to $k \!=\! 2^{m-1}k_0$, where $m$ is the index of the dense block, and $k_0$ is a constant. This way of setting the growth rate does not introduce any additional hyper-parameters. The ``increasing growth rate'' (IGR) strategy places a larger proportion of parameters in the later layers of the model. This increases the computational efficiency substantially but may decrease the parameter efficiency in some cases.
Depending on the specific hardware limitations it may be advantageous to trade-off one for the other~\cite{iandola2016squeezenet}.

\paragraph{Fully dense connectivity.}
To encourage feature re-use even more than the original DenseNet architecture does already, we connect input layers to \emph{all} subsequent layers in the network, even if these layers are located in different dense blocks (see \figurename~\ref{fig:fully_densenet}). As dense blocks have different feature resolutions, we  downsample feature maps with higher resolutions when we use them as inputs into lower-resolution layers using average pooling.

\section{Experiments}
We evaluate \condense{}s on the CIFAR-10, CIFAR-100 \cite{cifar}, and the ImageNet (ILSVRC 2012; \cite{deng2009imagenet}) image-classification datasets.
The models and code reproducing our experiments are publicly available at
{\small{\url{https://github.com/ShichenLiu/CondenseNet}}}.

\paragraph{Datasets.}
The CIFAR-10 and CIFAR-100 datasets consist of RGB images of size 32$\times$32 pixels, corresponding to 10 and 100 classes, respectively. Both datasets contain 50,000 training images and 10,000 test images.
We use a standard data-augmentation scheme \cite{netinnet, fitnet, dsn, allcnn, highway, stochastic, fractalnet}, in which the images are zero-padded with 4 pixels on each side, randomly cropped to produce 32$\times$32 images, and horizontally mirrored with probability $0.5$.

The ImageNet dataset comprises 1000 visual classes, and contains a total of 1.2 million training images and 50,000 validation images. We adopt the data-augmentation scheme of \cite{resnet} at training time, and perform a rescaling to $256 \times 256$ followed by a $224\times 224$ center crop at test time before feeding the input image into the networks.

\begin{figure}
  \centering
  \includegraphics[width=0.48 \textwidth]{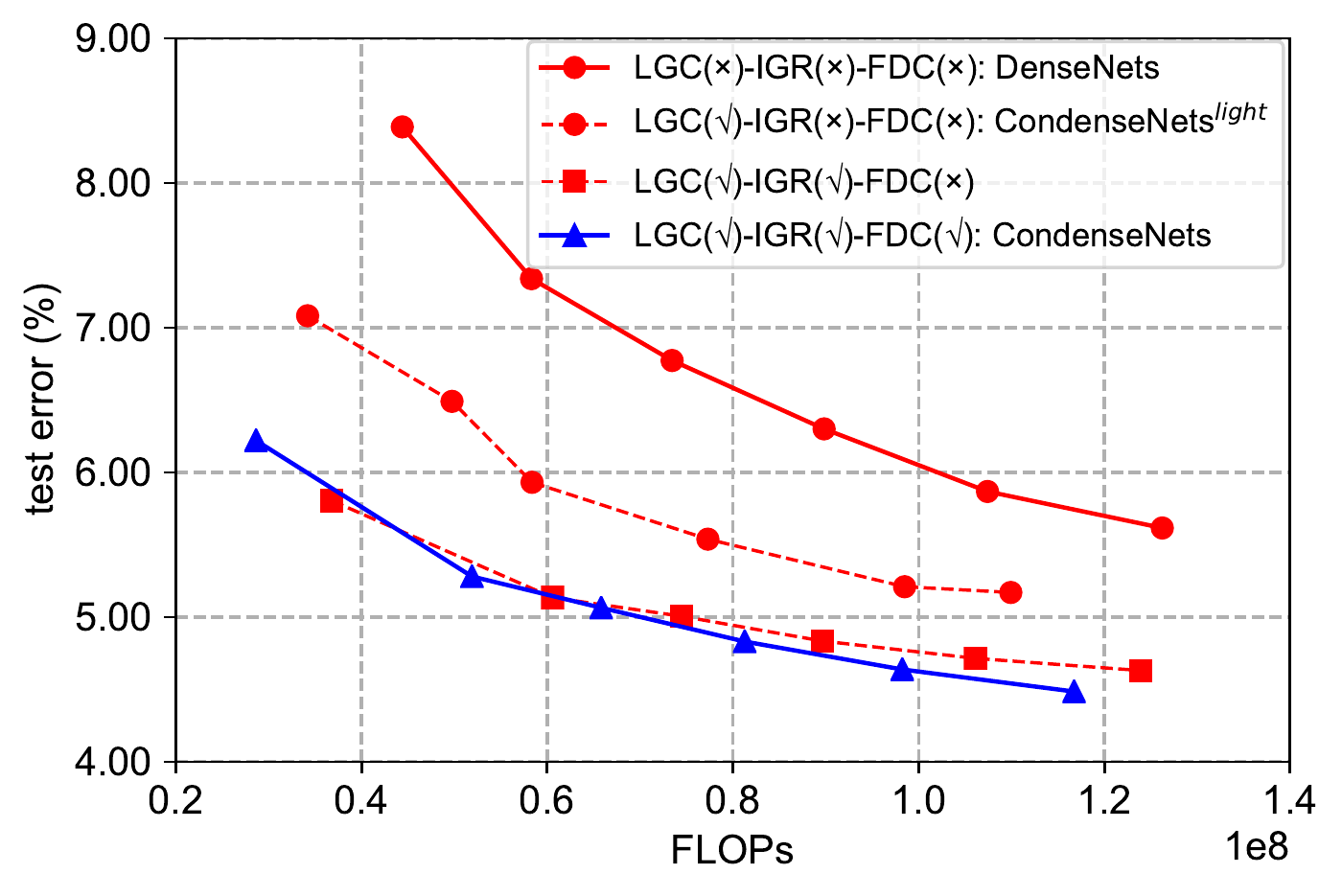}\\
  \caption{Ablation study on CIFAR-10 to investigate the efficiency gains obtained by the various components of \condense{}.}
  \label{fig:param_vs_err_pp}
\vspace{-2ex}
\end{figure}


\subsection{Results on CIFAR}
\vspace{-1ex}
\label{subsec:cifar}
We first perform a set of experiments on CIFAR-10 and CIFAR-100 to validate the effectiveness of \methodname{}s and the proposed \condense{} architecture.

\paragraph{Model configurations.}
Unless otherwise specified, we use the following network configurations in all experiments on the CIFAR datasets. The standard DenseNet has a constant growth rate of $k\!=\!12$ following \cite{huang2017densely}; our proposed architecture uses growth rates $k_0\!\in\!\{8,\! 16,\! 32\}$ to ensure that the growth rate is divisable by the number of groups. The \methodname{} is only applied to the first convolutional layer (with filter size $1\!\times\! 1$, see \figurename~\ref{fig:structure_compare}) of each basic layer, with a condensation factor of $C \!=\! 4$, \emph{i.e.}, 75\% of filter weights are gradually pruned during training with a step of 25\%.  The $3\!\times\! 3$ convolutional layers are replaced by standard group convolution (without applying \methodname) with four groups. Following  \cite{zhang2017shufflenet,zhang2017interleaved}, we permute the output channels of the first $1\!\times\! 1$ learned group convolutional layer, such that the features generated by each of its groups are evenly used by all the groups of the subsequent $3\times 3$ group convolutional layer .

\paragraph{Training details.}
We train all models with stochastic gradient descent (SGD) using similar optimization hyper-parameters as in \cite{resnet, huang2017densely}. Specifically, we adopt Nesterov momentum with a momentum weight of 0.9 without dampening, and use a weight decay of $10^{-4}$. All models are trained with mini-batch size 64 for 300 epochs, unless otherwise specified. We use a cosine shape learning rate which starts from 0.1 and gradually reduces to 0.
Dropout~\cite{dropout} with a drop rate of $0.1$ was applied to train \condense{}s with $>\!3$ million parameters (shown in Table~\ref{tab:cifar}).

\paragraph{Component analysis.}
\figurename~\ref{fig:param_vs_err_pp} compares the computational efficiency gains obtained by each component of CondenseNet: learned group convolution (LGR), exponentially increasing learning rate (IGR), full dense connectivity (FDC).
Specifically, the figure plots the test error as a function of the number of FLOPs (\emph{i.e.}, multiply-addition operations).
The large gap between the two red curves with dot markers shows that \methodname{} significantly improves the efficiency of our models. Compared to DenseNets, \condense$^{light}$ only requires half the number of FLOPs to achieve comparable accuracy.
Further, we observe that the exponentially increasing growth rate, yields even further efficiency. Full dense connectivity does not boost the efficiency significantly on CIFAR-10, but there does appear to be a trend that as models getting larger, full connectivity starts to help. We opt to include this architecture change in the \condense{} model, as it does lead to substantial improvements on ImageNet (see later).


\paragraph{Comparison with state-of-the-art efficient CNNs.}
In Table~\ref{tab:cifar}, we show the results of experiments comparing a 160-layer \condense$^\emph{{light}}$ and a 182-layer \condense{} with alternative state-of-the-art CNN architectures.
Following \cite{zoph2017learning}, our models were trained for 600 epochs. From the results, we observe that \condense{} requires approximately $8\times$ fewer parameters and FLOPs to achieve a comparable accuracy to DenseNet-190. \condense{} seems to be less parameter-efficient than \condense$^\emph{{light}}$, but is more compute-efficient. Somewhat surprisingly, our \condense$^\emph{{light}}$ model performs on par with the NASNet-A, an architecture that was obtained using an automated search procedure over $20,000$ candidate architectures composed of a rich set of components, and is thus carefully tuned on the CIFAR-10 dataset \cite{zoph2017learning}. Moreover, \condense{} (or \condense$^\emph{{light}}$)
does not use depth-wise separable convolutions, and only use simple convolutional filters with size $1\!\times\!1$ and $3\!\times\!3$. It may be possible to include \condense{} as a meta-architecture in the procedure of \cite{zoph2017learning} to obtain even more efficient networks.

\begin{table}[]
\addtolength{\tabcolsep}{0pt}
\centering
\small
\resizebox{0.48\textwidth}{!}{
\begin{tabular}{l|rrll}
\hline
\textbf{Model}               					& \textbf{Params}	& \textbf{FLOPs}			& \textbf{C-10}		& \textbf{C-100} \\ \hline
ResNet-1001\cite{identity-mappings}     & 16.1M		& 2,357M		& 4.62		& 22.71	\\
Stochastic-Depth-1202\cite{stochastic}       	& 19.4M		& 2,840M		& 4.91		& -	\\
Wide-ResNet-28\cite{wide}       		& 36.5M    	& 5,248M        & 4.00   	& 19.25     \\
ResNeXt-29 \cite{resnext}          		& 68.1M    	& 10,704M       & 3.58   	& 17.31     \\
DenseNet-190\cite{huang2017densely}		& 25.6M    	& 9,388M		& 3.46  	& 17.18     \\
NASNet-A$^*$\cite{zoph2017learning}		& 3.3M     	& -             & 3.41    	& -         \\ \hline
\condense$^\emph{\text{light}}$-160$^*$     					& 3.1M  	& 1,084M 		& 3.46		& 17.55	\\
\condense-182$^*$   				& 4.2M     	& 513M	        & 3.76 		& 18.47     \\
\hline
\end{tabular}
}
\vspace{2pt}
\caption{Comparison of classification error rate (\%) with other convolutional networks on the CIFAR-10(C-10) and CIFAR-100(C-100) datasets. * indicates models that are trained with cosine shape learning rate for 600 epochs.}
\label{tab:cifar}
\end{table}

\paragraph{Comparison with existing pruning techniques.}
In Table~\ref{tab:compare-prune}, we compare our \condense{}s and \condense s$^\emph{{light}}$ with models that are obtained by state-of-the-art \emph{filter-level} weight pruning techniques \cite{li2016pruning,liu2017learning,he2017channel}. The results show that, in general, \condense{} is about $3\times$ more efficient in terms of FLOPs than ResNets or DenseNets pruned by the method introduced in \cite{liu2017learning}. The advantage over the other pruning techniques is even more pronounced. We also report the results for \condense$^\emph{{light}}$ in the second last row of Table~\ref{tab:compare-prune}. It uses only half the number of parameters to achieve comparable performance as the most competitive baseline, the 40-layer DenseNet described by \cite{liu2017learning}.

\begin{table}[]
\addtolength{\tabcolsep}{0pt}
\centering
\small
\resizebox{0.48\textwidth}{!}{
\begin{tabular}{l|rrrr}
\hline
\textbf{Model}  & \textbf{FLOPs} & \textbf{Params} & \textbf{C-10} & \textbf{C-100} \\
\hline
VGG-16-pruned \cite{li2016pruning} 				& 206M 	&  5.40M   	&   6.60	& 25.28 \\
VGG-19-pruned \cite{liu2017learning} 			& 195M 	&  2.30M   	&   6.20 	&  - \\
VGG-19-pruned \cite{liu2017learning} 			& 250M 	&  5.00M   	&   -	 	& 26.52 \\
\hline
ResNet-56-pruned \cite{he2017channel}			& 62M	&  		&   8.20	& - \\
ResNet-56-pruned \cite{li2016pruning}			& 90M 	&  0.73M	&	6.94	& - \\
ResNet-110-pruned \cite{li2016pruning}			& 213M 	&  1.68M   	&   6.45 	& - \\
ResNet-164-B-pruned \cite{liu2017learning}	& 124M 	&  1.21M  	&   5.27		& {23.91}  \\
\hline
DenseNet-40-pruned \cite{liu2017learning}	& 190M 	&  0.66M   	&   5.19		& 25.28 \\
\hline
\condense$^\emph{\text{light}}$-94							& 122M 	&  0.33M  	&   {5.00} & 24.08 \\
\condense-86						& 65M 	&  0.52M  	&   {5.00} & {23.64} \\
\hline
\end{tabular}
}
\vspace{2.0pt}
\caption{Comparison of classification error rate (\%) on CIFAR-10 (C-10) and CIFAR-100 (C-100) with state-of-the-art filter-level weight pruning methods.}
\vspace{-1ex}
\label{tab:compare-prune}
\end{table}

\renewcommand{\arraystretch}{1.0}
\begin{table}[!t]
\addtolength{\tabcolsep}{5pt}
\centering
\small
{

\begin{tabular}{c|c}
\hline
\textbf{\condense}       	& \textbf{Feature map size}				   \\ \hline
\cross{3} Conv (stride $2$) 			& \cross{112}                  \\ \hline
\conv{4}{8}	        				& \cross{112}                  \\ \hline
\cross{2} average pool, stride 2 	& \cross{56}                   \\ \hline
\conv{6}{16}        				& \cross{56}                   \\ \hline
\cross{2} average pool, stride 2 	& \cross{28}                   \\ \hline
\conv{8}{32}        				& \cross{28}                   \\ \hline
\cross{2} average pool, stride 2 	& \cross{14}                   \\ \hline
\conv{10}{64}       				& \cross{14}                   \\ \hline
\cross{2} average pool, stride 2 	& \cross{7}                    \\ \hline
\conv{8}{128}        				& \cross{7}                    \\ \hline
\cross{7} global average pool 		& \cross{1}                    \\ \hline
1000-dim fully-connected, softmax 	&   						   \\ \hline
\end{tabular}
}
\vspace{1 ex}
\caption{\condense{} architectures for ImageNet.}
\label{tab:structure-imagenet}
\vspace{-3 ex}
\end{table}

\subsection{Results on ImageNet}
\vspace{-1ex}
In a second set of experiments, we test \condense{} on the ImageNet dataset.

\paragraph{Model configurations.}
Detailed network configurations are shown in Table~\ref{tab:structure-imagenet}. To reduce the number of parameters, we prune 50\% of weights from the fully connected (FC) layer at epoch 60 in a way similar to the \methodname{}, but with $G\!=\!1$ (as the FC layer could not be split into multiple groups) and $C\!=\!2$. Similar to prior studies on MobileNets and ShuffleNets, we focus on training relatively small models that require less than 600 million FLOPs to perform inference on a single image.

\paragraph{Training details.}
We train all models using stochastic gradient descent (SGD) with a batch size of 256. As before, we adopt Nesterov momentum with a momentum weight of 0.9 without dampening, and a weight decay of $10^{-4}$. All models are trained for 120 epochs, with a cosine shape learning rate which starts from 0.1 and gradually reduces to 0. We use group lasso regularization in all experiments on ImageNet; the regularization parameter is set to $10^{-5}$.

\paragraph{Comparison with state-of-the-art efficient CNNs.}
Table~\ref{tab:imagenet} shows the results of \condense{}s and several state-of-the-art, efficient models on the ImageNet dataset. We observe that a \condense{}  with 274 million FLOPs obtains a 29.0\% Top-1 error, which is comparable to the accuracy achieved by MobileNets and ShuffleNets that require twice as much compute. A \condense{} with 529 million FLOPs produces to a 3\% absolute reduction in top-1 error compared to a MobileNet and a ShuffleNet of comparable size. Our \condense{} even achieves a the same accuracy with slightly fewer FLOPs and parameters than the most competitive NASNet-A, despite the fact that we only trained a very small number of models (as opposed to the study that lead to the NASNet-A model).

\paragraph{Actual inference time.}
Table~\ref{tab:wall-time} shows the actual inference time on an ARM processor for different models. The wall-time to inference an image sized at $224\times 224$ is highly correlated with the number of FLOPs of the model. Compared to the recently proposed MobileNet, our \condense{} ($G\!=\!C\!=\!8$) with 274 million FLOPs inferences an image $2\times$ faster, while without sacrificing accuracy.

\renewcommand{\arraystretch}{1.1}
\begin{table}[!t]
\centering
\small
\addtolength{\tabcolsep}{-3pt}
\begin{tabular}{l|rrrr}
\hline
\textbf{Model}     & \textbf{FLOPs} & \textbf{Params} & \textbf{Top-1} & \textbf{Top-5} \\ \hline
Inception V1 \cite{szegedy2015going}       & 1,448M                     & 6.6M                 & 30.2                     & 10.1                     \\
1.0 MobileNet-224 \cite{howard2017mobilenets}  & 569M                       & 4.2M                 & 29.4                     & 10.5                     \\
ShuffleNet 2x  \cite{zhang2017shufflenet}    & 524M                       & 5.3M                   & 29.1                     & 10.2                     \\
NASNet-A (N=4) \cite{zoph2017learning}    & 564M                       & 5.3M                 & {26.0}   & 8.4                     \\
NASNet-B (N=4) \cite{zoph2017learning}    & 488M                       & 5.3M                 & {27.2}   & 8.7                     \\
NASNet-C (N=3)  \cite{zoph2017learning}   & 558M                       & 4.9M                 &{27.5}   & 9.0                     \\\hline
\condense{} ($G\!=\!C\!=\!8$)	   &  {274M}      &  {2.9M}	        & {{29.0}}            &{{10.0}}  \\
\condense{} ($G\!=\!C\!=\!4$)   & 529M		       & 4.8M	        & 26.2	          & {8.3}            \\ \hline
\end{tabular}
\vspace{2ex}
\caption{Comparison of \emph{Top-1} and \emph{Top-5} classification error rate (\%) with other state-of-the-art compact models on ImageNet.}
\label{tab:imagenet}
\vspace{-2ex}
\end{table}

\renewcommand{\arraystretch}{1.1}
\begin{table}[!t]
\centering
\small
\addtolength{\tabcolsep}{-3pt}
\begin{tabular}{l|rrcc}
\hline
\textbf{Model}     										& \textbf{FLOPs} 	& \textbf{Top-1} 	& \textbf{Time(s)}	 	\\ \hline
VGG-16      								    & 15,300M	& 28.5		& 354		\\
ResNet-18       								& 1,818M	& 30.2		& 8.14		\\
1.0 MobileNet-224 \cite{howard2017mobilenets}  	& 569M		& 29.4		& 1.96		\\
\condense{}  ($G\!=\!C\!=\!4$)					& 529M 		& 26.2		& 1.89 		\\
\condense{}  ($G\!=\!C\!=\!8$)					& 274M 		& 29.0		& 0.99 		\\ \hline
\end{tabular}
\vspace{2ex}
\caption{Actual inference time of different models on an ARM processor. All models are trained on ImageNet, and accept input with resolution 224$\times$224.}
\label{tab:wall-time}
\vspace{-2ex}
\end{table}

\begin{figure*}[ht]
	\centering
	\includegraphics[width=0.33 \textwidth]{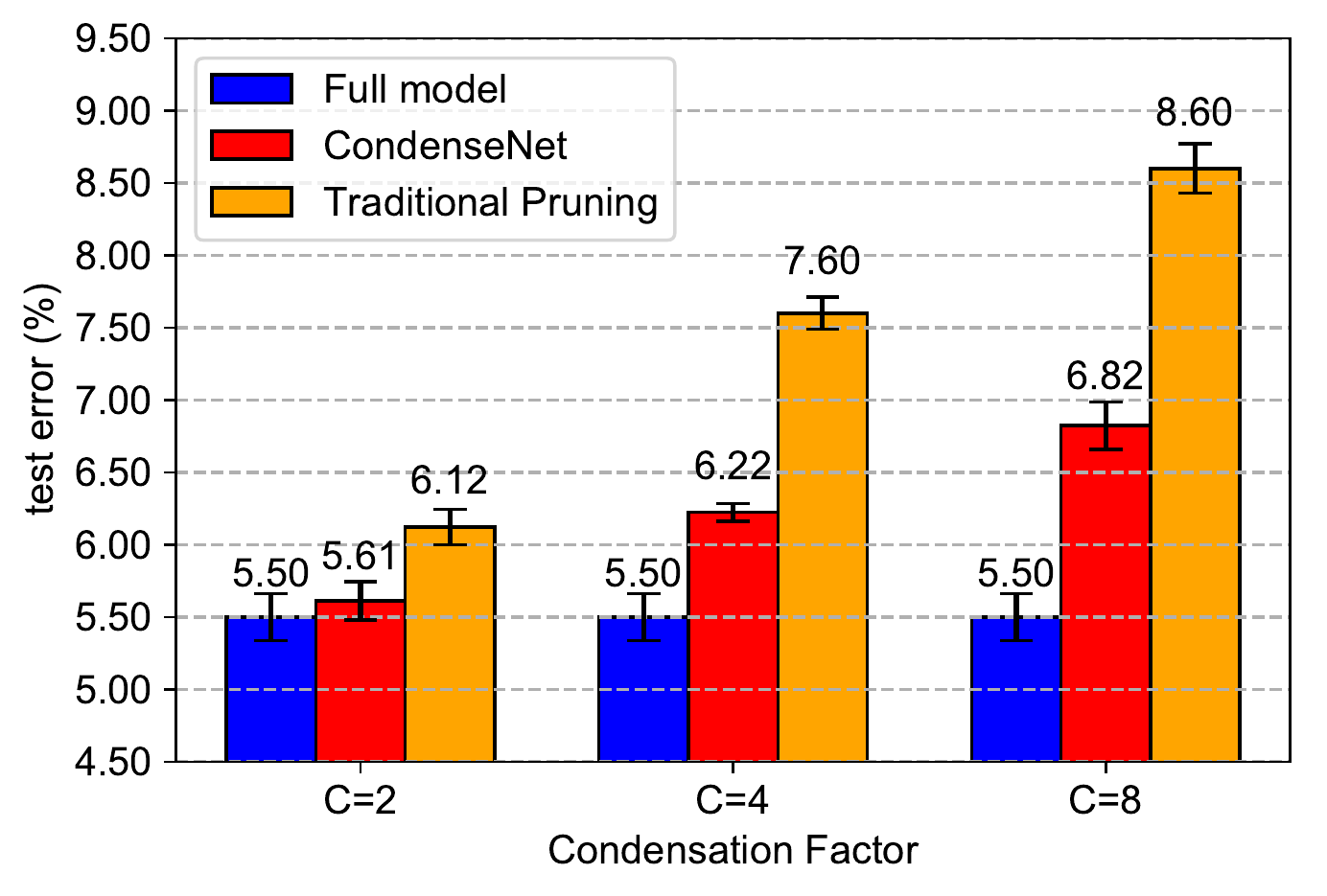}
    \includegraphics[width=0.33 \textwidth]{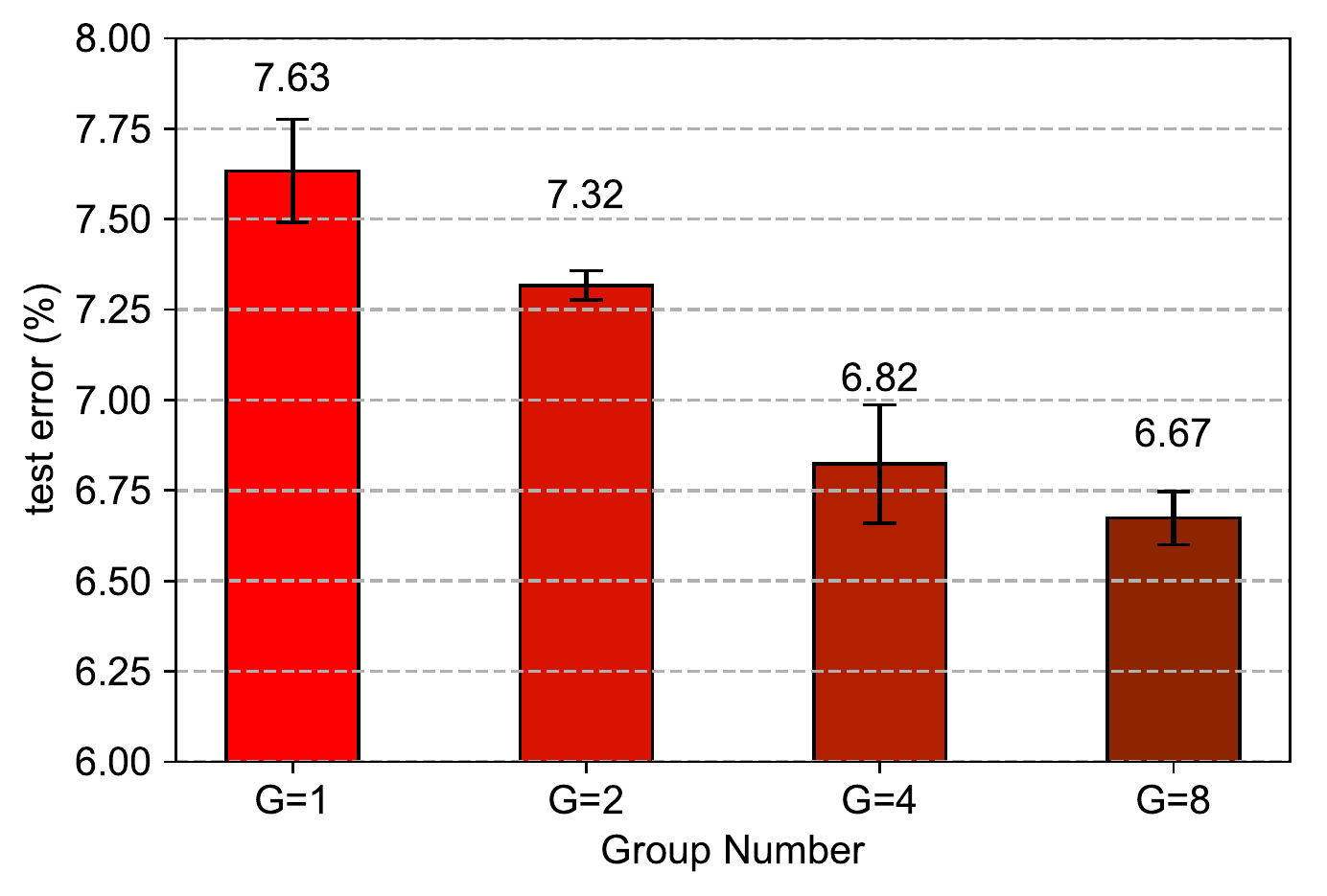}
	\includegraphics[width=0.33 \textwidth]{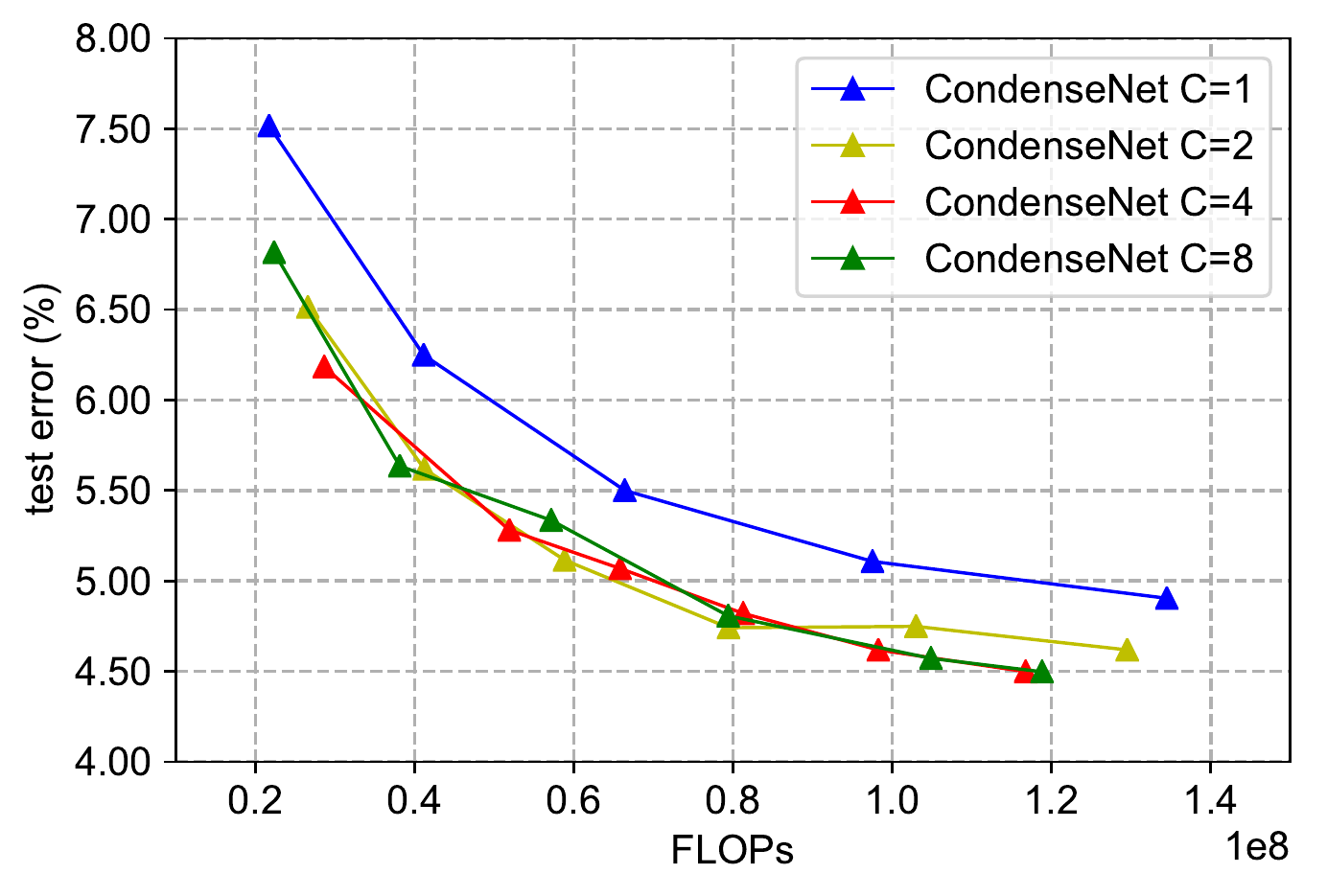}
	\caption{Classification error rate (\%) on CIFAR-10. \emph{Left}: Comparison between our condense method with traditional pruning approach, under varying condensation factors. \emph{Middle}: \condense{}s with different number of groups for the $1\!\times\! 1$ \methodname{}. All the models have the same number of parameters. \emph{Right}: \condense{}s with different condensation factors. }
	\label{fig:vary-G-C}
\vspace{-2 ex}
\end{figure*}

\subsection{Ablation Study}
\vspace{-1ex}
We perform an ablation study on CIFAR-10 in which we investigate the effect of (1) the pruning strategy, (2) the number of groups, and (3) the condensation factor. We also investigate the stability of our weight pruning procedure.

\paragraph{Pruning strategy.}
The left panel of \figurename~\ref{fig:vary-G-C} compares our \emph{on-the-fly} pruning method with the more common approach of pruning weights of \emph{fully converged} models. We use a DenseNet with $50$ layers as the basis for this experiment. We implement a ``traditional'' pruning method in which the weights are pruned in the same way as in as in \condense{}s, but the pruning is only done once after training has completed (for 300 epochs). Following~\cite{liu2017learning}, we fine-tune the resulting sparsely connected network for another 300 epochs with the same cosine shape learning rate that we use for training \condense{}s. We compare the traditional pruning approach with the \condense{} approach, setting the number of groups $G$ is set to $4$. In both settings, we vary the condensation factor $C$ between $2$ and $8$.

The results in \figurename~\ref{fig:vary-G-C} show that pruning weights gradually during training outperforms pruning weights on fully trained models. Moreover, gradual weight pruning reduces the training time: the ``traditional pruning'' models were trained for $600$ epochs, whereas the \condense{}s were trained for $300$ epochs. The results also show that removing 50\% the weights (by setting $C\!=\!2$) from the $1 \!\times\! 1$ convolutional layers in a DenseNet incurs hardly any loss in accuracy.

\paragraph{Number of groups.} In the middle panel of \figurename~\ref{fig:vary-G-C}, we compare four \condense{}s with exactly the same network architecture, but a number of groups, $G$, that varies between $1$ and $8$. We fix the condensation factor, $C$, to 8 for all the models, which implies all models have the same number of parameters after training has completed. In \condense{}s with a single group, we discard entire filters in the same way that is common in filter-pruning techniques \cite{li2016pruning,liu2017learning}. The results presented in the figure demonstrate that test errors tends to decrease as the number of groups increases. This result is in line with our analysis in Section \ref{sec:method}, in particular, it suggests that grouping filters gives the training algorithm more flexibility to remove redundant weights.


\begin{figure}[t]
	\centering
	\includegraphics[width=0.48 \textwidth]{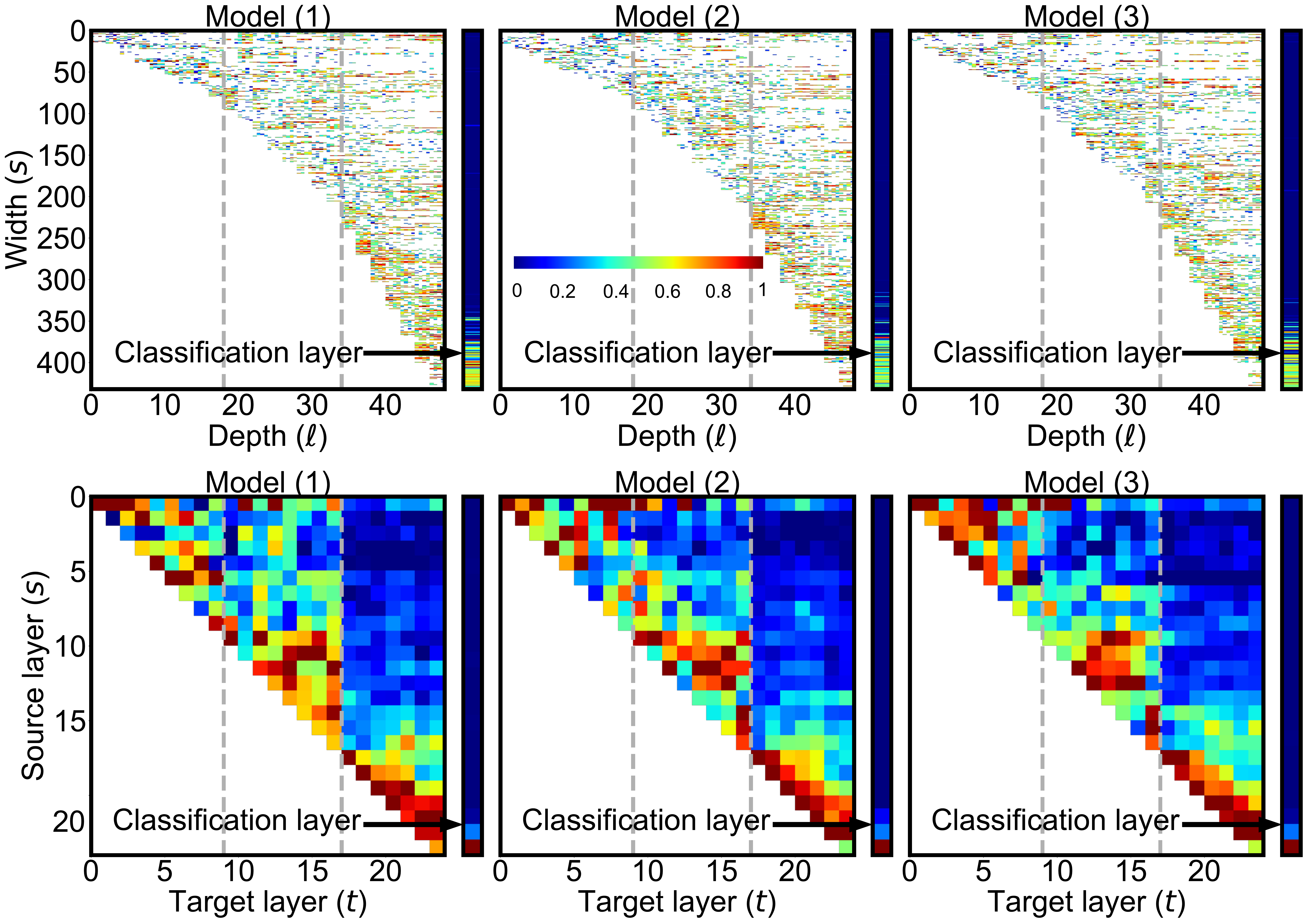}
	\caption{Norm of weights between layers of a CIFAR-10 \condense{} per filter group (\emph{top}) and per filter block (\emph{bottom}). The three columns correspond to independent training runs.}
	\label{fig:learned-weights-stablity}
\vspace{-2ex}
\end{figure}

\paragraph{Effect of the condensation factor.} In the right panel of \figurename~\ref{fig:vary-G-C}, we compare \condense{}s with varying condensation factors. Specifically, we set the condensation factor $C$ to 1, 2, 4, or 8; this corresponds to removing 0\%, 50\%, 75\%, or 87.5\% of the weights from each of the $1\!\times\! 1$ convolutional layers, respectively. A condensation factor $C\!=\!1$ corresponds to a baseline model without weight pruning. The number of groups, $G$, is set to 4 for all the networks.
The results show that a condensation factors $C$ larger than 1 consistently lead to improved efficiency, which underlines the effectiveness of our method. Interestingly, models with condensation factors 2, 4 and 8 perform comparably in terms of classification error as a function of FLOPs. This suggests that whilst pruning more weights yields smaller models, it also leads to a proportional loss in accuracy.


%
%
%

%

\paragraph{Stability.} As our method removes redundant weights in early stages of the training process, a natural question is whether this will introduce extra variance into the training. Does early pruning remove some of the weights simply because they were initialized with small values?

To investigate this question, \figurename~\ref{fig:learned-weights-stablity} visualizes the learned weights and connections for three independently trained \condense{}s on CIFAR-10 (using different random seeds). The top row shows detailed weight strengths (averaged absolute value of non-pruned weights) between a filter group of a certain layer (corresponding to a column in the figure) and an input feature map (corresponding to a row in the figure). For each layer there are four filter groups (consecutive columns). A white pixel in the top-right corner indicates that a particular input feature was pruned by that layer and group.
Following \cite{huang2017densely}, the bottom row of \figurename~{fig:learned-weights-stablity} shows the overall connection strength between two layers in the condensed network. The vertical bars correspond to the linear classification layer on top of the \condense{}. The gray vertical dotted lines correspond to pooling layers that decrease the feature resolution.

The results in the figure suggest that while there are differences in learned connectivity at the filter-group level (top row), the overall information flow between layers (bottom row) is similar for all three models. This suggests that the three training runs learn similar global connectivity patterns, despite starting from different random initializations. Later layers tend to prefer more recently generated features, do however utilize some features from very early layers.

\vspace{-1ex}
\section{Conclusion}
\vspace{-1ex}
In this paper, we introduced \condense{}: an efficient convolutional network architecture that encourages feature re-use via dense connectivity and prunes filters associated with superfluous feature re-use via \methodname{}s. To make inference efficient, the pruned network can be converted into a network with regular group convolutions, which are implemented efficiently in most deep-learning libraries. Our pruning method is simple to implement, and adds only limited computational costs to the training process. In our experiments, \condense{}s outperform recently proposed MobileNets and ShuffleNets in terms of computational efficiency at the same accuracy level. \condense{} even slightly outperforms a network architecture that was discovered by empirically trying tens of thousands of convolutional network architectures, and with a much simpler structure.

\paragraph{Acknowledgements.}
The authors are supported in part by grants from the National Science Foundation ( III-1525919, IIS-1550179, IIS-1618134, S\&AS 1724282, and CCF-1740822), the Office of Naval Research DOD (N00014-17-1-2175), and the Bill and Melinda Gates Foundation. We are thankful for generous support by SAP America Inc. We also thank Xu Zou, Weijia Chen, Danlu Chen for helpful discussions.

{\small
\bibliographystyle{ieee}
\bibliography{citations}}

\end{document}